\documentclass[11pt]{article}

\usepackage{acl}
\usepackage{times}
\usepackage{latexsym}
\usepackage[T1]{fontenc}
\usepackage[utf8]{inputenc}
\usepackage{microtype}
\usepackage{inconsolata}
\usepackage{graphicx}
\usepackage{amsmath}
\usepackage{booktabs}
\usepackage{array}
\usepackage{enumitem}
\usepackage{xspace}
\usepackage{float}

\graphicspath{{figures/}{appendix/}}

\newcommand{\benchmark}{LDU-Bench\xspace}
\newcommand{\lcs}{LCS\xspace}
\newcommand{\dicu}{DICU\xspace}
\newcommand{\cmd}{CMD\xspace}
\newcommand{\efs}{EFS\xspace}

\title{\benchmark: Multimodal LLM Evaluation for Lithography Defect Understanding under Layout-Varying Circuit Backgrounds}
\author{
  Huanglong Ji \And
  Botong Zhao \And
  Shujing Lv \And
  Yue Lv\thanks{Corresponding author.}
}

\begin{document}
\maketitle

\begin{abstract}
Multimodal large language models have demonstrated strong defect recognition capability in industrial anomaly detection. However, in lithography review, merely determining whether an image contains a defect is insufficient for engineering inspection; models must also understand defect morphology, spatial location, and the potential causes supported by visible evidence. To this end, this paper proposes LDU-Bench, a multi-task multimodal benchmark for lithography defect understanding. Constructed from real lithography and integrated-circuit review images, LDU-Bench decomposes the review workflow into four independent tasks: defect triage, morphology recognition, coarse localization, and image-conditioned cause analysis. It systematically evaluates models using task-level metrics, diagnostic readouts, and the Lithography Closure Score (LCS). Experimental results show that although existing MLLMs can perform defect triage relatively reliably, this ability does not stably transfer to downstream review stages. Morphology alignment, effective localization, and evidence-to-cause mapping remain the major bottlenecks. Further diagnostics indicate that this capability break is not a fluctuation of a single metric, but reflects insufficient structured understanding across semantic levels. Overall, LDU-Bench provides a quantifiable and diagnostic unified platform for evaluating the usability, failure points, and capability boundaries of industrial MLLMs in lithography review chains.
\end{abstract}

\section{Introduction}

Existing semiconductor and lithography defect inspection work largely frames the problem as classification, detection, localization, or segmentation \citep{dehaerne2025semreview,deridder2023semicenternet,shin2016cnn,yang2017lithography,liao2022lithography,dey2022deep}. These methods can output classes, bounding boxes, or segmentation masks, but they do not directly provide morphology descriptions, spatial explanations, or cause evidence that can be used in downstream human review.
\begin{figure}[H]
\centering
\includegraphics[width=\columnwidth]{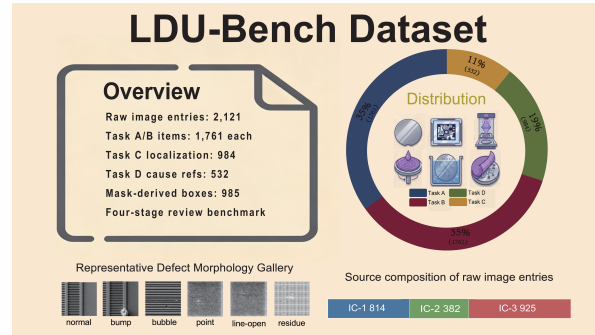}
\caption{Dataset and task split. \benchmark uses lithography IC-SEM review images with image-level labels, mask-derived spatial annotations, and reviewed cause references.}
\label{fig:dataset}
\end{figure}

Recent industrial anomaly detection work has introduced vision-language interfaces. Compared with labels, boxes, or masks, natural language can express morphology, location, and cause cues, and is therefore closer to the judgment required in review. MVTec AD, VisA, Real-IAD, and MVTec LOCO standardize anomaly detection and segmentation tasks \citep{bergmann2019mvtec,bergmann2022beyond,zou2022spot,wang2024realiad}. Representative industrial anomaly detection methods further study patch-distribution modeling, reconstruction-based discrimination, memory-based retrieval, and efficient student-teacher detection \citep{defard2021padim,zavrtanik2021draem,roth2022patchcore,batzner2024efficientad}. Building on image-text alignment from CLIP \citep{radford2021clip}, WinCLIP, ALFA, AnomalyCLIP, and AnomalyGPT further adapt vision-language models to anomaly recognition, segmentation, and interactive localization \citep{jeong2023winclip,zhu2024alfa,zhou2024anomalyclip,gu2024anomalygpt}. MMAD organizes industrial anomaly detection as multimodal question answering and systematically evaluates MLLMs \citep{jiang2025mmad}. However, existing work mainly targets general industrial objects. Its evaluation objectives remain anomaly discovery, segmentation, or QA accuracy, and do not specifically test whether strong defect triage in lithography implies downstream morphology recognition, coarse localization, and image-conditioned cause analysis.

Lithography review places stronger requirements on this capability chain. The appearance of many industrial inspection datasets is relatively stable, whereas the normal background of circuit images varies with design layer, local routing, and pattern density. Defects are often embedded in dense periodic or semi-periodic structures. Therefore, review understanding requires more than detecting a local defect. A model must organize local visual morphology, spatial location, and possible cause into an auditable information chain. Evaluating only whether a defect is found cannot determine whether the model has downstream review understanding.

Therefore, we propose LDU-Bench, a multi-task multimodal benchmark for lithography defect understanding.
It is constructed from real lithography and integrated-circuit review images and divides the review process into four independently evaluated tasks: defect triage, morphology recognition, coarse localization, and image-conditioned cause analysis.
LDU-Bench uses deterministic scorers, diagnostic readouts, and LCS as an overall review-chain summary.
This design directly tests whether defect-triage performance is sufficient to indicate downstream review understanding.
The results show that high triage scores do not imply stable performance in later review stages.
For the evaluated MLLMs, morphology grounding, localization utility, and evidence-to-cause mapping remain the main bottlenecks.

\section{Method}

\subsection{Overall Framework}

\begin{figure*}[t]
\centering
\includegraphics[width=0.95\textwidth]{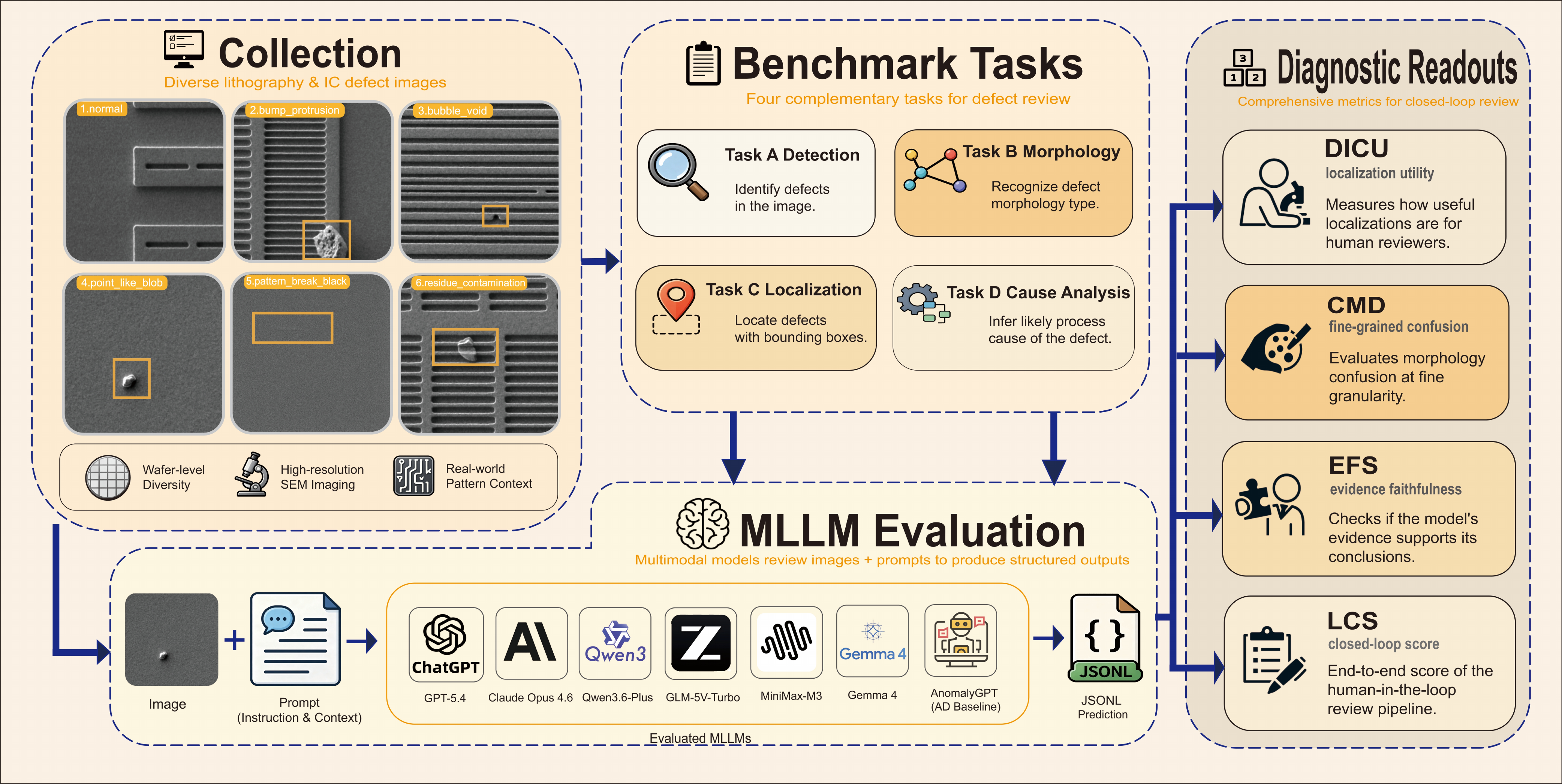}
\caption{Overall workflow of \benchmark. The benchmark decomposes lithography review into defect triage, morphology recognition, coarse localization, and image-conditioned cause analysis, while also evaluating task-level performance and diagnostic readouts.}
\label{fig:workflow}
\end{figure*}

The core of \benchmark is to test whether models can map defect evidence to reviewable morphology, location, and image-supported cause information. Therefore, we organize lithography defect understanding into four independent tasks: defect triage, morphology recognition, coarse localization, and image-conditioned cause analysis. These tasks follow the order of review information needs, but they are evaluated independently. This design avoids cascade error and directly tests whether defect triage can stably extend to downstream review understanding.

This design allows \benchmark to answer two questions. First, how does a model perform at each review stage. Second, whether the model's review ability can remain continuous across stages, rather than being strong only at one stage. In addition to task-level metrics, we introduce \lcs as an overall summary to describe the model's completion of the four review stages. By combining task-level scores and \lcs, \benchmark can compare overall performance while exposing where the review chain becomes weak. Figure~\ref{fig:workflow} shows the structure of the \benchmark dataset.

\subsection{Dataset Construction}

We construct LDU-Bench from IC-SEM lithography review images, covering multiple semiconductor process stages, including BEOL, DEP, and DPR.

The dataset is organized with a unified manifest. This manifest integrates image-level annotations, spatial regions, and expert-reviewed cause labels using a consistent schema.

This design supports multi-stage evaluation of defect understanding. Each image is associated within a unified framework with triage labels, morphology categories, spatial annotations, and cause-level explanations.

\subsection{Benchmark Tasks and Metrics}

The scoring goal of LDU-Bench is not to produce a single overall ranking, but to answer a more specific question: at which review stage does a model begin to become unreliable. Therefore, we first define the main metric for each of the four tasks, then use diagnostic analyses to explain the sources of low scores, and finally use LCS to provide an overall score.

Task A performs defect triage on 1,761 images, requiring the model to make a binary decision between \texttt{good} and \texttt{defect}. This corresponds to the initial screening stage in industrial review workflows. Task B uses the same image set and requires predicting a fine-grained morphology label from \texttt{normal} plus 11 defect categories. Task B can be viewed as a fine-grained extension of Task A, requiring the model to further match the detection result with the corresponding morphology label after defect detection.

Task C evaluates spatial grounding on 984 mask-annotated samples, where the model must output an \texttt{xyxy} bounding box. This setting follows the common evaluation paradigm of grounded vision-language tasks \citep{peng2023kosmos,chen2023shikra,you2023ferret} and simplifies pixel-level masks into bounding box expressions, allowing the evaluation to focus on coarse-grained spatial localization ability.

Task D uses 532 expert-reviewed curated samples to construct an image-conditioned cause reasoning task, requiring the model to predict a cause label while also giving a brief rationale.

In evaluation, Task A and Task B use macro-F1. This metric is more robust to class imbalance in the morphology distribution, and accuracy is used as a secondary metric. Task C uses Defect Intersection-Coverage Utility (DICU), which jointly accounts for overlap quality (IoU) and region coverage (GT coverage):
\[
\mathrm{DICU}_i =
\frac{2\cdot \mathrm{IoU}_i\cdot \mathrm{GTcov}_i}
{\mathrm{IoU}_i+\mathrm{GTcov}_i}.
\]

Task D adopts a structured rubric-based evaluation. Exact matches or alias matches for the cause are scored as 1, semantically related matches within the same supergroup are scored as 0.5, and all other cases are scored as 0. The final score combines semantic correctness and keyword-level evidence alignment. The weights balance semantic correctness and evidence consistency, with a higher weight on semantic matching to reflect the primary objective of cause attribution:
\[
S_D = 0.7\cdot \mathrm{Sem}_i + 0.3\cdot \mathrm{KeyF1}_i.
\]

To further analyze sources of model failure beyond task-level performance, we introduce two diagnostic metrics. CMD compares label-only and definition-guided prompts on confusable defect pairs to distinguish whether morphology errors arise from visual ambiguity or label understanding difficulty. EFS evaluates the visual faithfulness of explanations in Task D, measuring the consistency between generated rationales and observable visual evidence, and identifying inference bias driven by hallucination or non-visual evidence \citep{li2023pope,guan2024hallusionbench}.

We use a power-mean form of LCS to summarize overall performance:
\[
\mathrm{LCS}_{p}
=
\left(
\frac{S_A^{p}+S_B^{p}+S_C^{p}+S_D^{p}}{4}
\right)^{1/p},\quad p>0,
\]
where $S_A$, $S_B$, $S_C$, and $S_D$ denote macro-F1, macro-F1, mean DICU, and mean Task D rubric score, respectively. The main results use $p=0.5$. The model ranking remains stable under $p \in \{0.25, 0.5, 0.75, 1.0\}$.

\section{Experiments}

\subsection{Model Evaluation Setup}

Our evaluation covers three types of models. The first type is commercial general MLLMs: GPT-5.4 (OpenAI, 2026), Claude Opus 4.6 (Anthropic, 2026), Qwen3.6-Plus (Alibaba, 2026), GLM-5V-Turbo (Zhipu AI, 2026), and MiniMax-M3 (MiniMax, 2026). We keep one representative model for each provider and keep model release time as consistent as possible, so that a single provider does not dominate the table with multiple versions. The second type is open-weight general MLLMs. We use locally deployed Gemma 4 31B-it (Google DeepMind, 2026) as a reproducible reference. The third type is a local anomaly-detection baseline. We deploy and obtain AnomalyGPT results on Task A and Task C to compare defect triage and coarse localization.

\subsection{\benchmark Dataset and Metrics}

The evaluation in LDU-Bench contains four tasks. Task A and Task B each contain 1,761 images. Task C contains 984 images with spatial annotations. Task D contains 532 reviewed image-conditioned cause references.

One complete A/B/C/D evaluation contains 5,038 task instances. Expanded to six full-chain models, this corresponds to 30,228 model-task evaluation units. With the additional evaluation of AnomalyGPT on Task A and Task C, the overall evaluation covers 32,973 task-level instances.

Task A and Task B use macro-F1 as performance metrics. Task C uses DICU to evaluate spatial grounding ability. Task D uses a deterministic rubric score to evaluate cause reasoning quality. Meanwhile, the diagnostic metric CMD is used to analyze error sources in Task A/B by comparing label-only and definition-guided prompts on confusable defect pairs, distinguishing whether morphology errors arise from visual ambiguity or label understanding difficulty. The diagnostic metric EFS is sampled from Task D outputs to evaluate the visual faithfulness of image-grounded explanations, namely whether the generated rationale depends on visible visual evidence rather than reasoning bias driven by language priors or hallucinated information.

\subsection{Inference and Scoring Protocol}

To make model differences mainly come from task ability, we control randomness during inference and scoring. For all controllable decoding interfaces, we use deterministic settings. Commercial APIs that expose a temperature parameter are all set to 0, and the output length is uniformly limited to 1024 to prevent overly long model outputs.

We evaluate Gemma 4 and AnomalyGPT under a unified batch inference setup, using fixed seeds to ensure reproducibility and fair comparison. Scoring uses the same task prompts, output schemas, and parsing rules. To ensure objective and reproducible scoring, the main task scores never use an LLM judge. Except for the human \efs diagnostic, all A/B/C/D metrics are generated by deterministic scorers. Across the six full-chain models, the overall task-level output validity rate is 99.88\%; detailed validity results are reported in Appendix Table X.

\subsection{Main Results and Diagnostic Findings}

Table~\ref{tab:main_results} summarizes the main results on the four \benchmark tasks. Columns A/B report macro-F1, column C reports \dicu, column D reports the deterministic rubric score, and the \lcs column reports the four-task closure score. AnomalyGPT is included as an anomaly-detection baseline and contains only A/C results.

\begin{table*}[t]
\centering
\small
\setlength{\tabcolsep}{4pt}
\begin{tabular}{lrrrrrl}
\toprule
Model & Task A & Task B & Task C & Task D & \lcs$_{0.5}$ & Output validity \\
 & macro-F1 & macro-F1 & \dicu & score & & \\
\midrule
GPT-5.4         & \textbf{0.932} & 0.201 & \textbf{0.492} & 0.409 & \textbf{0.474} & \textbf{100.00\%} \\
GLM-5V-Turbo    & 0.884 & \textbf{0.408} & 0.353 & 0.297 & 0.462 & \textbf{100.00\%} \\
MiniMax-M3      & 0.909 & 0.239 & 0.338 & \textbf{0.395} & 0.439 & \textbf{100.00\%} \\
Qwen3.6-Plus    & 0.856 & 0.249 & 0.269 & 0.376 & 0.408 & 99.96\% \\
Claude Opus 4.6 & 0.876 & 0.266 & 0.254 & 0.255 & 0.378 & \textbf{100.00\%} \\
Gemma 4 31B-it  & 0.504 & 0.060 & 0.046 & 0.080 & 0.132 & 99.34\% \\
\midrule
AnomalyGPT & 0.759 & -- & 0.384 & -- & -- & 100.00\% on A/C \\
\bottomrule
\end{tabular}
\caption{Main results on \benchmark. Bold values indicate the best performance in each column. AnomalyGPT is an AD baseline and is not evaluated on the full review chain.}
\label{tab:main_results}
\end{table*}

The most important finding in Table~\ref{tab:main_results} is not the model ranking, but that defect-triage ability does not stably transfer to downstream review understanding tasks. The five commercial MLLMs achieve high F1 on Task A, showing that current models already have some defect-triage ability. However, this advantage does not carry over to Task B/C/D. In other words, models can usually judge whether an image contains a defect, but they cannot stably answer what morphology the defect has, where it is located, and which possible cause is supported by visible evidence.

This gap first appears in Task B. GPT-5.4 obtains the highest scores on Task A, Task C, and Task D, but drops to 0.201 on morphology recognition. GLM-5V-Turbo obtains the highest Task B score of 0.408, but does not maintain the same advantage in other review stages. This indicates that Task B is not a natural extension of Task A. A model that can complete defect triage does not necessarily map local visual morphology stably to the controlled morphology labels in \benchmark. Therefore, morphology naming is the first bottleneck between defect detection and review understanding.

The results of Task C and Task D further show that this performance gap is not limited to the level of label naming, but extends across two deeper capability dimensions: localization and cause reasoning. On spatial grounding and cause explanation tasks, the upper bound of current commercial MLLMs remains limited. The highest DICU in Task C is only 0.492, and the highest rubric score in Task D is only 0.409. This indicates that even the strongest model struggles to simultaneously provide stable coarse localization and reliable image-evidence-based attribution.

In the comparison across model types, AnomalyGPT outperforms all commercial MLLMs except GPT-5.4 on Task C, indicating that specialized industrial anomaly detection models still have advantages on coarse localization. However, this model does not support morphology recognition or cause analysis, and therefore cannot form a complete review reasoning chain. By contrast, Gemma 4 31B-it, as the open-weight baseline model, lags clearly behind closed-source models on all four tasks, showing that current general open-weight foundation MLLMs still have a substantial capability gap in this structured lithography review task.

\begin{figure*}[t]
\centering
\includegraphics[width=0.96\textwidth]{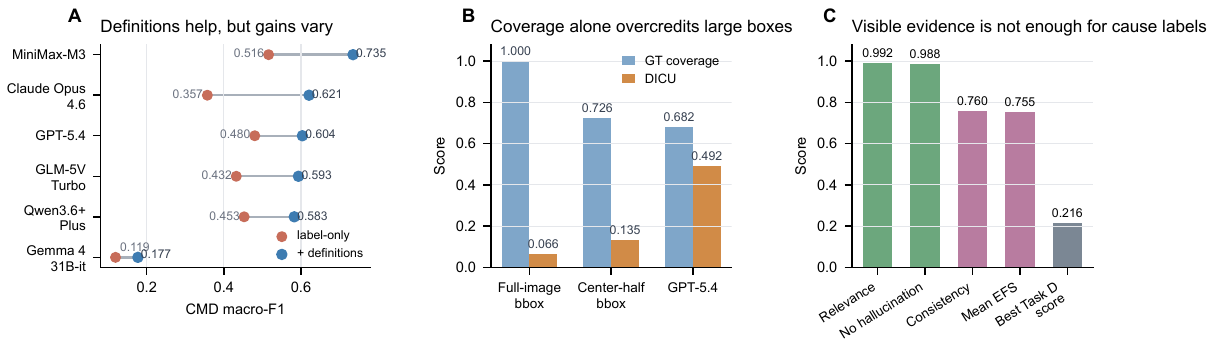}
\caption{Diagnostic readouts for morphology grounding, localization utility, and evidence-to-cause mapping. Panel (A) reports \cmd macro-F1 under label-only and definition-guided settings. Panel (B) compares GT coverage and \dicu for geometric boxes and GPT-5.4. Panel (C) reports \efs pilot scores and the best Task D score on the same subset.}
\label{fig:diagnostics}
\end{figure*}

To explain the failure causes behind these scores, Figure~\ref{fig:diagnostics} reports diagnostics from three angles: morphology grounding, localization utility, and evidence-to-cause mapping. Figure~\ref{fig:diagnostics}A shows that some morphology recognition errors can be mitigated by short definitions. GPT-5.4 improves from 0.530 accuracy / 0.480 macro-F1 under label-only prompting to 0.630 / 0.604 under definition-guided prompting. Other commercial models also reach 0.583 to 0.661 macro-F1 in definition-guided \cmd. This shows that lower Task B scores cannot be fully attributed to visual discrimination failure. Because fab-specific morphology labels differ across fabs in annotation standards and defect definitions, label semantics are inconsistent across sources, which affects the model's ability to maintain stable semantic alignment.

Figure~\ref{fig:diagnostics}B shows that effective localization cannot be judged only by defect coverage. Large geometric boxes can easily obtain high GT coverage. For example, the full-image box has GT coverage of 1.000, but its mIoU is only 0.0387 and its \dicu is only 0.0663. The center-half box still has 0.7257 GT coverage, but its \dicu is only 0.1352. This means that overly large prediction boxes cover the defect while also covering a large amount of irrelevant background. Real model localizations are better than these simple geometric methods, but still do not reach reliable review quality. GPT-5.4 obtains the highest \dicu, 0.4917, while other commercial models mainly fall between 0.25 and 0.35. Therefore, the bottleneck of Task C is not whether a model can roughly point to an abnormal region, but whether it can provide a compact and useful box for review.

Figure~\ref{fig:diagnostics}C further analyzes evidence faithfulness in Task D outputs. Among 250 manually scored responses, mean relevance is 0.992, mean consistency is 0.760, hallucination rate is 1.2\%, and mean \efs is about 0.755. This means that models usually write visible anomaly cues and rarely explicitly cite image-external information. However, the Task D score on the same subset remains low. The best model reaches only 0.216. This shows that the presence of visible evidence is not equivalent to correct cause attribution. A model may capture local anomaly cues, but still fail to map them to the controlled cause reference. Therefore, this paper mainly uses \efs to audit whether Task D explanations have image evidence, rather than as a model-ranking metric.

In addition to individual diagnostics, \lcs reveals systematic differences in model capability at the overall review-chain level. Table~\ref{tab:main_results} shows that although commercial models generally perform similarly on Task A and have small variance, the final \lcs ranking shows clear stratification (GPT-5.4: 0.474; GLM-5V-Turbo: 0.462; the remaining models decrease in order). This indicates that overall performance differences mainly do not come from triage ability, but are dominated by the accumulation of capabilities in downstream review stages.

Further analysis shows that GPT-5.4's advantage comes from stable performance on Task C and Task D, while GLM-5V-Turbo narrows the gap at the morphology modeling stage through its relative advantage on Task B. By contrast, other models show capability degradation in multiple review stages, causing their \lcs values to be gradually separated. Therefore, \lcs is not a linear combination of single-task performance, but an overall characterization of the ability to continuously complete the full chain from defect triage to structured review reasoning.

Combining the main results and diagnostic analyses, the major limitation of current MLLMs is not the absence of defect perception, but the difficulty of maintaining stable ability as task semantic level increases. Although models perform relatively stably on the basic defect-triage task (Task A), performance shows a systematic downward trend when tasks extend to higher-level structured understanding, such as morphology grounding, spatial localization (Task C), and evidence-to-cause reasoning (Task D). This phenomenon indicates that existing MLLMs are more biased toward low-level visual discrimination and lack consistent generalization in structured understanding and causal reasoning across semantic levels.

\subsection{Reliability Analysis}
\label{sec:reliability}

This section further tests whether the evaluation protocol provides stable and interpretable signals. We mainly analyze the annotation reliability of morphology labels.

For Task B, we conduct an independent blind-labeling experiment on 100 samples to test whether the frozen morphology taxonomy in \benchmark is sufficient to support the main evaluation. The result reaches 79.0\% exact agreement and Cohen's $\kappa=0.727$ \citep{cohen1960coefficient}, indicating high annotator consistency. Further inspection shows that disagreements mainly occur between visually adjacent categories, such as \texttt{bump\_protrusion} and \texttt{buried\_anomaly}. This shows that Task B is not an arbitrary subjective classification task, but has an auditable label basis.

\section{Conclusion}

Task-level scores, diagnostic experiments, and \lcs together give a consistent conclusion: current MLLMs show a systematic capability-structure gap in lithography review tasks. This is precisely where the value of \benchmark lies. Its value is not in building a more complex defect recognition benchmark, but in abstracting the review process into a quantifiable and diagnostic multi-level capability evaluation framework, thereby characterizing model behavior differences across semantic levels. The experimental results show that although existing commercial MLLMs already have some anomaly-triage ability, this ability does not naturally extend to stable structured review understanding. This phenomenon indicates that current models still have a systematic break in the continuous modeling process from visual perception to semantic attribution, rather than a lack of ability in a single module.

Therefore, the essence of the problem is not limited to a bottleneck in a specific task, but is an overall alignment problem across visual perception, spatial representation, and causal reasoning.

\section*{Limitations}

LDU-Bench has several controlled scope boundaries that arise from building a publicly releasable lithography-review benchmark. First, its morphology labels follow a fixed benchmark taxonomy designed to support consistent evaluation across all tested models. Independent blind labeling shows that this taxonomy provides an auditable basis for Task B. Nevertheless, lithography review practices may vary across fabs, equipment conditions, and annotation protocols. We therefore use Task B to evaluate whether models can align visual defect morphology with a consistent set of review labels, while leaving broader cross-fab taxonomy harmonization to future work. Second, Task D focuses on image-conditioned cause attribution. Since process logs, equipment states, recipe parameters, and historical inspection records are often sensitive and difficult to release in a public benchmark, we currently evaluate whether visible defect cues can be matched to reviewed cause references, instead of attempting full process-level causal diagnosis. Future extensions can incorporate fab-specific taxonomies, process metadata, temporal records, and human-in-the-loop feedback under industrial collaboration, enabling evaluation closer to closed-loop lithography diagnosis.

\bibliography{references}

@article{shin2016cnn,
  title = {CNN Based Lithography Hotspot Detection},
  author = {Shin, Moojoon and Lee, Jee-Hyong},
  journal = {International Journal of Fuzzy Logic and Intelligent Systems},
  volume = {16},
  number = {3},
  pages = {208--215},
  year = {2016}
}

@inproceedings{yang2017lithography,
  title = {Lithography Hotspot Detection: From Shallow to Deep Learning},
  author = {Yang, Haoyu and Lin, Yajun and Yu, Bei and Young, Evangeline F. Y.},
  booktitle = {2017 30th IEEE International System-on-Chip Conference (SOCC)},
  pages = {233--238},
  year = {2017},
  organization = {IEEE},
  doi = {10.1109/SOCC.2017.8226047}
}

@article{liao2022lithography,
  title = {Lithography Hotspot Detection Method Based on Transfer Learning Using Pre-Trained Deep Convolutional Neural Network},
  author = {Liao, Lufeng and Li, Sikun and Che, Yongqiang and Shi, Weijie and Wang, Xiangzhao},
  journal = {Applied Sciences},
  volume = {12},
  number = {4},
  pages = {2192},
  year = {2022},
  doi = {10.3390/app12042192}
}

@inproceedings{dey2022deep,
  title = {Deep Learning Based Defect Classification and Detection in SEM Images: A Mask R-CNN Approach},
  author = {Dey, Bappaditya and Dehaerne, Enrique and Halder, Sandip and Leray, Philippe and Bayoumi, Magdy A.},
  booktitle = {Metrology, Inspection, and Process Control XXXVI},
  volume = {PC12053},
  pages = {PC120530K},
  year = {2022},
  organization = {SPIE},
  doi = {10.1117/12.2618178}
}

@article{dehaerne2025semreview,
  title = {Scanning Electron Microscopy-Based Automatic Defect Inspection for Semiconductor Manufacturing: A Systematic Review},
  author = {Dehaerne, Enrique and Dey, Bappaditya and Blanco, Victor and Davis, Jesse},
  journal = {Journal of Micro/Nanopatterning, Materials, and Metrology},
  volume = {24},
  number = {2},
  pages = {020901},
  year = {2025},
  doi = {10.1117/1.JMM.24.2.020901}
}

@inproceedings{deridder2023semicenternet,
  title = {SEMI-CenterNet: A Machine Learning Facilitated Approach for Semiconductor Defect Inspection},
  author = {De Ridder, Vic and Dey, Bappaditya and Dehaerne, Enrique and Halder, Sandip and De Gendt, Stefan and Van Waeyenberge, Bartel},
  booktitle = {38th European Mask and Lithography Conference (EMLC 2023)},
  volume = {12802},
  pages = {128020M},
  year = {2023},
  organization = {SPIE},
  doi = {10.1117/12.2675570}
}

@inproceedings{bergmann2019mvtec,
  title = {MVTec AD---A Comprehensive Real-World Dataset for Unsupervised Anomaly Detection},
  author = {Bergmann, Paul and Fauser, Michael and Sattlegger, David and Steger, Carsten},
  booktitle = {Proceedings of the IEEE/CVF Conference on Computer Vision and Pattern Recognition},
  pages = {9592--9600},
  year = {2019},
  doi = {10.1109/CVPR.2019.00982}
}

@article{bergmann2022beyond,
  title = {Beyond Dents and Scratches: Logical Constraints in Unsupervised Anomaly Detection and Localization},
  author = {Bergmann, Paul and Batzner, Kilian and Fauser, Michael and Sattlegger, David and Steger, Carsten},
  journal = {International Journal of Computer Vision},
  volume = {130},
  pages = {947--969},
  year = {2022},
  doi = {10.1007/s11263-022-01578-9}
}

@article{zou2022spot,
  title = {SPot-the-Difference Self-Supervised Pre-training for Anomaly Detection and Segmentation},
  author = {Zou, Yang and Jeong, Jongheon and Pemula, Latha and Zhang, Dongqing and Dabeer, Onkar},
  journal = {arXiv preprint arXiv:2207.14315},
  year = {2022}
}

@inproceedings{wang2024realiad,
  title = {Real-IAD: A Real-World Multi-View Dataset for Benchmarking Versatile Industrial Anomaly Detection},
  author = {Wang, Chengjie and Zhu, Wenbing and Gao, Bin-Bin and Gan, Zhenye and Zhang, Jianning and Gu, Zhihao and Qian, Shuguang and Chen, Mingang and Ma, Lizhuang},
  booktitle = {Proceedings of the IEEE/CVF Conference on Computer Vision and Pattern Recognition},
  year = {2024}
}

@inproceedings{defard2021padim,
  title = {PaDiM: A Patch Distribution Modeling Framework for Anomaly Detection and Localization},
  author = {Defard, Thomas and Setkov, Aleksandr and Loesch, Angelique and Audigier, Romaric},
  booktitle = {Pattern Recognition. ICPR International Workshops and Challenges},
  pages = {475--489},
  year = {2021},
  publisher = {Springer},
  doi = {10.1007/978-3-030-68799-1_35}
}

@inproceedings{zavrtanik2021draem,
  title = {DRAEM: A Discriminatively Trained Reconstruction Embedding for Surface Anomaly Detection},
  author = {Zavrtanik, Vitjan and Kristan, Matej and Sko{\v{c}}aj, Danijel},
  booktitle = {Proceedings of the IEEE/CVF International Conference on Computer Vision},
  pages = {8330--8339},
  year = {2021},
  doi = {10.1109/ICCV48922.2021.00822}
}

@inproceedings{roth2022patchcore,
  title = {Towards Total Recall in Industrial Anomaly Detection},
  author = {Roth, Karsten and Pemula, Latha and Zepeda, Joaquin and Sch{\"o}lkopf, Bernhard and Brox, Thomas and Gehler, Peter},
  booktitle = {Proceedings of the IEEE/CVF Conference on Computer Vision and Pattern Recognition},
  pages = {14318--14328},
  year = {2022},
  doi = {10.1109/CVPR52688.2022.01392}
}

@inproceedings{batzner2024efficientad,
  title = {EfficientAD: Accurate Visual Anomaly Detection at Millisecond-Level Latencies},
  author = {Batzner, Kilian and Heckler, Lars and K{\"o}nig, Rebecca},
  booktitle = {Proceedings of the IEEE/CVF Winter Conference on Applications of Computer Vision},
  pages = {127--137},
  year = {2024},
  doi = {10.1109/WACV57701.2024.00020}
}

@inproceedings{radford2021clip,
  title = {Learning Transferable Visual Models from Natural Language Supervision},
  author = {Radford, Alec and Kim, Jong Wook and Hallacy, Chris and Ramesh, Aditya and Goh, Gabriel and Agarwal, Sandhini and Sastry, Girish and Askell, Amanda and Mishkin, Pamela and Clark, Jack and Krueger, Gretchen and Sutskever, Ilya},
  booktitle = {Proceedings of the 38th International Conference on Machine Learning},
  volume = {139},
  pages = {8748--8763},
  year = {2021},
  publisher = {PMLR}
}

@inproceedings{jeong2023winclip,
  title = {WinCLIP: Zero-/Few-Shot Anomaly Classification and Segmentation},
  author = {Jeong, Jongheon and Zou, Yang and Kim, Taewan and Zhang, Dongqing and Ravichandran, Avinash and Dabeer, Onkar},
  booktitle = {Proceedings of the IEEE/CVF Conference on Computer Vision and Pattern Recognition},
  pages = {19606--19616},
  year = {2023}
}

@inproceedings{zhou2024anomalyclip,
  title = {AnomalyCLIP: Object-Agnostic Prompt Learning for Zero-Shot Anomaly Detection},
  author = {Zhou, Qihang and Pang, Guansong and Tian, Yu and He, Shibo and Chen, Jiming},
  booktitle = {International Conference on Learning Representations},
  year = {2024}
}

@inproceedings{gu2024anomalygpt,
  title = {AnomalyGPT: Detecting Industrial Anomalies Using Large Vision-Language Models},
  author = {Gu, Zhaopeng and Zhu, Bingke and Zhu, Guibo and Chen, Yingying and Tang, Ming and Wang, Jinqiao},
  booktitle = {Proceedings of the AAAI Conference on Artificial Intelligence},
  volume = {38},
  number = {3},
  pages = {1932--1940},
  year = {2024},
  doi = {10.1609/aaai.v38i3.27963}
}

@inproceedings{zhu2024alfa,
  title = {Do LLMs Understand Visual Anomalies? Uncovering LLM's Capabilities in Zero-shot Anomaly Detection},
  author = {Zhu, Jiaqi and Cai, Shaofeng and Deng, Fang and Ooi, Beng Chin and Wu, Junran},
  booktitle = {Proceedings of the 32nd ACM International Conference on Multimedia},
  pages = {48--57},
  year = {2024},
  doi = {10.1145/3664647.3681190}
}

@inproceedings{jiang2025mmad,
  title = {MMAD: A Comprehensive Benchmark for Multimodal Large Language Models in Industrial Anomaly Detection},
  author = {Jiang, Xi and Li, Jian and Deng, Hanqiu and Liu, Yong and Gao, Bin-Bin and Zhou, Yifeng and Li, Jialin and Wang, Chengjie and Zheng, Feng},
  booktitle = {International Conference on Learning Representations},
  year = {2025}
}

@article{peng2023kosmos,
  title = {Kosmos-2: Grounding Multimodal Large Language Models to the World},
  author = {Peng, Zhiliang and Wang, Wenhui and Dong, Li and Hao, Yaru and Huang, Shaohan and Ma, Shuming and Wei, Furu},
  journal = {arXiv preprint arXiv:2306.14824},
  year = {2023}
}

@article{chen2023shikra,
  title = {Shikra: Unleashing Multimodal LLM's Referential Dialogue Magic},
  author = {Chen, Keqin and Zhang, Zhao and Zeng, Weili and Zhang, Richong and Zhu, Feng and Zhao, Rui},
  journal = {arXiv preprint arXiv:2306.15195},
  year = {2023}
}

@article{you2023ferret,
  title = {Ferret: Refer and Ground Anything Anywhere at Any Granularity},
  author = {You, Haoxuan and Zhang, Haotian and Gan, Zhe and Du, Xianzhi and Zhang, Bowen and Wang, Zirui and Cao, Liangliang and Chang, Shih-Fu and Yang, Yinfei},
  journal = {arXiv preprint arXiv:2310.07704},
  year = {2023}
}

@inproceedings{li2023pope,
  title = {Evaluating Object Hallucination in Large Vision-Language Models},
  author = {Li, Yifan and Du, Yifan and Zhou, Kun and Wang, Jinpeng and Zhao, Wayne Xin and Wen, Ji-Rong},
  booktitle = {Proceedings of the 2023 Conference on Empirical Methods in Natural Language Processing},
  pages = {292--305},
  year = {2023}
}

@inproceedings{guan2024hallusionbench,
  title = {HallusionBench: An Advanced Diagnostic Suite for Entangled Language Hallucination and Visual Illusion in Large Vision-Language Models},
  author = {Guan, Tianrui and Liu, Fuxiao and Wu, Xiyang and Xian, Ruiqi and Li, Zongxia and Liu, Xiaoyu and Wang, Xijun and Chen, Lichang and Huang, Furong and Yacoob, Yaser and Manocha, Dinesh and Zhou, Tianyi},
  booktitle = {Proceedings of the IEEE/CVF Conference on Computer Vision and Pattern Recognition},
  pages = {14375--14385},
  year = {2024}
}

@article{cohen1960coefficient,
  title = {A Coefficient of Agreement for Nominal Scales},
  author = {Cohen, Jacob},
  journal = {Educational and Psychological Measurement},
  volume = {20},
  number = {1},
  pages = {37--46},
  year = {1960},
  doi = {10.1177/001316446002000104}
}

\clearpage
\appendix
\section{Experimental Configuration and Reproducibility Details}
\label{app:experiment_configuration}

This appendix reports implementation settings that are omitted from the main text for space. All model runs use the same frozen task files, image inputs, prompt templates, output schemas, parsing rules, and deterministic scorers as the main evaluation. Commercial API models are evaluated with deterministic decoding whenever the interface exposes the relevant option. Local models are evaluated in batch mode with fixed seeds.

\subsection{Model and Runtime Settings}

\begin{table*}[t]
\centering
\small
\setlength{\tabcolsep}{3pt}
\begin{tabular}{p{0.16\textwidth}p{0.18\textwidth}p{0.21\textwidth}p{0.34\textwidth}}
\toprule
Group & Model & Runtime / adapter & Main configuration \\
\midrule
Commercial MLLM & GPT-5.4 & OpenAI Responses-compatible endpoint & Original image input with high-detail image setting from the adapter; two workers; one retry. Decoding controls are kept at the endpoint defaults when not exposed in the run metadata. \\
Commercial MLLM & Claude Opus 4.6 & Anthropic Messages-compatible endpoint & Original image encoded in the message payload; temperature $0$; maximum output length $512$ tokens; two workers; four retries; $2$ s sleep between retries; $600$ s timeout. \\
Commercial MLLM & Qwen3.6-Plus & OpenAI-compatible chat endpoint & High-detail image input; temperature $0$; maximum output length $1024$ tokens; four workers; six retries; $1$ s sleep between retries; $180$ s timeout. \\
Commercial MLLM & GLM-5V-Turbo & OpenAI-compatible chat endpoint & Original task image input; temperature $0$; maximum output length $1024$ tokens; four workers; four retries; $3$ s sleep between retries; $600$ s timeout. \\
Commercial MLLM & MiniMax-M3 & Anthropic Messages-compatible endpoint & Original image encoded in the message payload; temperature $0$; maximum output length $512$ tokens; one worker; three retries; $1$ s sleep between retries; $300$ s timeout. \\
Open-weight MLLM & Gemma 4 31B-it & Local HuggingFace VLM adapter & Checkpoint \texttt{models/gemma-4-31B-it}; processor-default image input; \texttt{max\_new\_tokens=512}; \texttt{do\_sample=false}; 4-bit BitsAndBytes NF4 quantization; \texttt{bfloat16} compute; \texttt{device\_map=auto}; batch inference on 8$\times$ NVIDIA A800 80GB GPUs. \\
Domain AD baseline & AnomalyGPT & Local anomaly-detection baseline & Evaluated on Task A and Task C only; task images as input; fixed seed \texttt{20260607}; mask-level AD-Seg summary uses seed \texttt{20260608} and map size $224\times224$; local batch evaluation on the same A800 cluster. \\
\bottomrule
\end{tabular}
\caption{Runtime and decoding settings used for the main evaluation. API keys and service credentials are not part of the appendix.}
\label{tab:appendix_runtime_settings}
\end{table*}

\subsection{Inference Controls and Scoring}

\begin{table*}[t]
\centering
\small
\setlength{\tabcolsep}{4pt}
\begin{tabular}{p{0.20\textwidth}p{0.70\textwidth}}
\toprule
Item & Setting \\
\midrule
Prompt and schema & Each task uses a frozen prompt template and a task-specific output schema. The prompts ask for structured answers rather than free-form prose whenever the metric requires deterministic parsing. \\
Image input & Evaluation uses the original task image files. API adapters encode images according to the provider interface. The local Gemma adapter uses the checkpoint processor defaults. \\
Decoding & Temperature is set to $0$ for API adapters that expose the field. Local Gemma inference uses \texttt{do\_sample=false}. Provider-side controls that are not exposed are left at the endpoint default and recorded in run metadata. \\
Invalid outputs & Invalid JSON, illegal labels, missing answers, and unparsable coordinates are not manually repaired or imputed. They remain in the scoring denominator and are counted as invalid or missing outputs. \\
Task A/B scoring & Classification outputs are scored with deterministic label normalization. Macro-F1 is the main metric for both tasks, and accuracy is retained as an auxiliary reading. \\
Task C scoring & Coordinates are parsed as \texttt{xyxy} boxes. Unparsable or illegal boxes receive zero localization utility. Valid boxes are scored by DICU, which combines IoU and ground-truth coverage. \\
Task D scoring & Cause labels and short rationales are scored by a frozen rubric with deterministic label, alias, supergroup, and keyword matching rules. No LLM judge is used for the main A/B/C/D metrics. \\
EFS diagnostic & Evidence-Faithfulness Score is a human diagnostic for Task D explanations. It is reported as exploratory analysis and is not included in the main ranking or LCS. \\
\bottomrule
\end{tabular}
\caption{Shared inference and scoring controls. These settings define the end-to-end evaluation protocol rather than model-specific hyperparameters.}
\label{tab:appendix_scoring_controls}
\end{table*}

\subsection{Supplemental Diagnostic Sampling}

Supplemental analyses use deterministic sampling so that the reported diagnostic sets can be regenerated. The CMD pair-generation script uses seed \texttt{20260602}. The Task D EFS diagnostic sampling script uses seed \texttt{20260604}. The random geometric localization baseline uses seed \texttt{20260608}. The Task B inter-annotator agreement review sheet uses seed \texttt{20260609}, and the Task D EFS verification sample uses seed \texttt{20260610}. These seeds only control diagnostic sampling or baseline construction; they do not change the frozen benchmark labels or task definitions.

\clearpage
\section{Output Validity and Parse Rates}
\label{app:validity_parse_rates}

Invalid JSON, illegal labels, missing answers, and unparsable coordinates are counted as invalid or missing outputs. They are not manually repaired or imputed, and they remain in the scoring denominator. A low parse rate should therefore be interpreted as lower end-to-end task reliability, because format following is part of whether a model can be used in a benchmarked review pipeline.

\begin{table*}[t]
\centering
\includegraphics[width=0.98\textwidth]{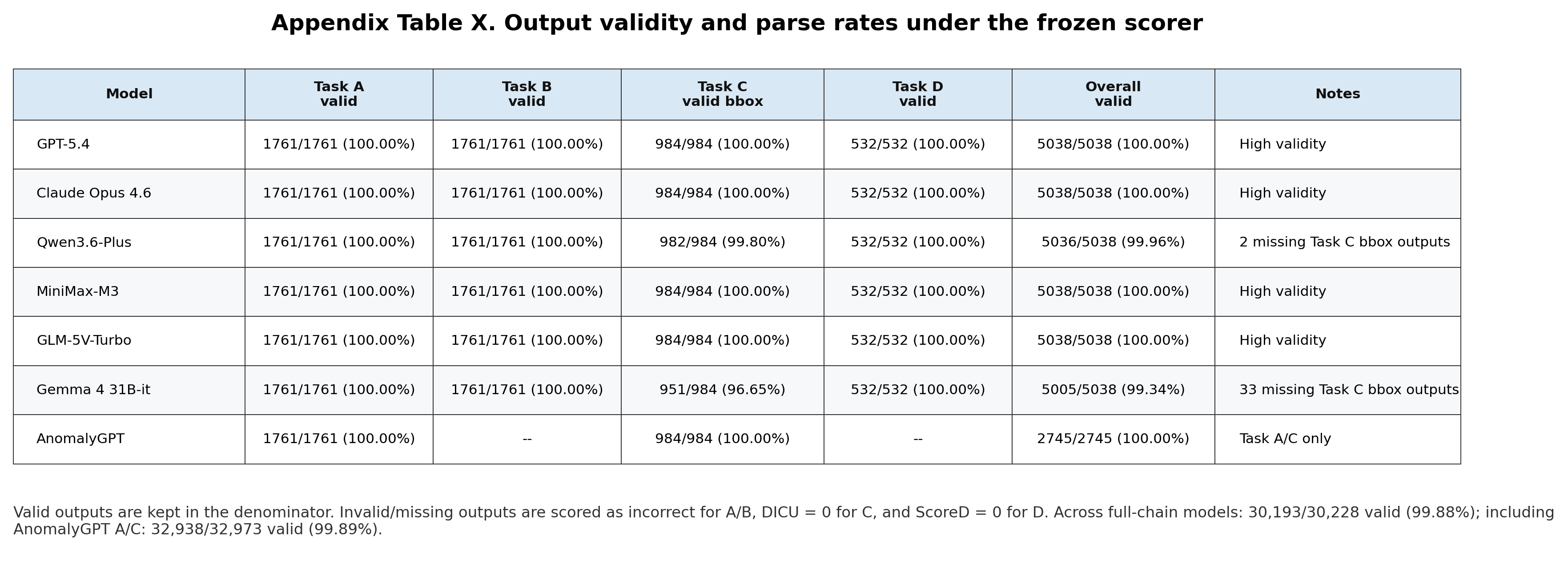}
\caption*{\textbf{Appendix Table X.} Output validity and parse rates under the frozen scorer.}
\label{tab:appendix_validity_parse_rates}
\end{table*}

\section{\lcs Parameter Sensitivity}
\label{app:lcs_sensitivity}

We test whether the choice of $p=0.5$ drives the model ranking. Holding the four task scores fixed, we compute \lcs for $p\in\{0.25,0.5,0.75,1.0\}$. Smaller $p$ values penalize weak stages more strongly, while $p=1.0$ is the arithmetic mean. The ranking is unchanged across all tested values.

\begin{table*}[t]
\centering
\includegraphics[width=0.98\textwidth]{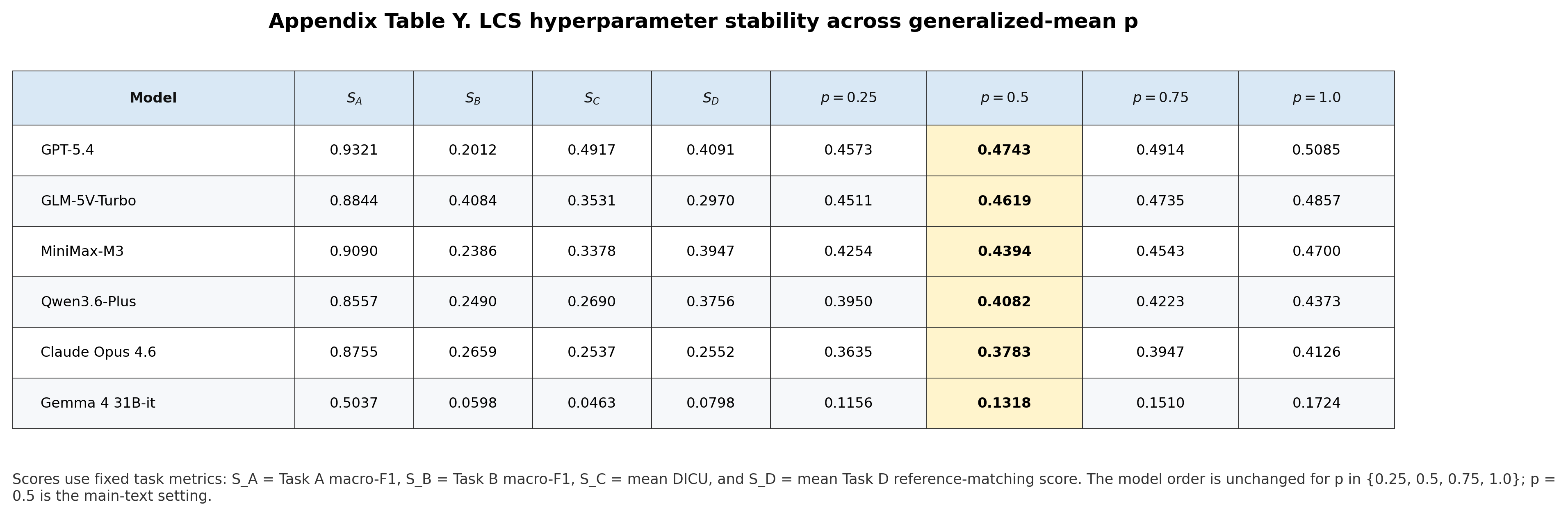}
\caption*{\textbf{Appendix Table Y.} LCS hyperparameter stability across generalized-mean $p$.}
\label{tab:appendix_lcs_p_sensitivity}
\end{table*}

\end{document}